\documentclass[journal]{IEEEtran}

\usepackage{moreverb,url}
\usepackage{color, colortbl}
\usepackage[labelfont=bf, hypcap=true,labelsep=space]{caption,subcaption}
\usepackage{booktabs}
\usepackage{array,multirow,graphicx} 
\usepackage{bbding} 
\usepackage{amsmath} 
\usepackage{amssymb}  
\usepackage{floatrow}
\usepackage{algorithm,algpseudocode}
\usepackage{caption}
\usepackage[hang]{footmisc}
\usepackage[bookmarks,bookmarksnumbered]{hyperref}
\hypersetup{colorlinks = true, linkcolor = blue, anchorcolor = red, citecolor = blue, filecolor = red, urlcolor = red, pdfauthor=author}
\usepackage{cleveref}
\usepackage{xcolor}
\usepackage[utf8]{inputenc}
\usepackage{cite}
\usepackage{siunitx}
\usepackage{footnote}
\usepackage{flushend}
\usepackage[bookmarks,bookmarksnumbered]{hyperref}

\definecolor{d_gray}{gray}{0.05}
\definecolor{l_gray}{gray}{0.95}
\definecolor{ll_gray}{gray}{0.97}

\begin{document}

\title{Introducing VaDA: Novel Image Segmentation Model for Maritime Object Segmentation Using New Dataset\\
}

\author{\IEEEauthorblockN{Yongjin Kim, Jinbum Park, Sanha Kang, Hanguen Kim} \\
\IEEEauthorblockA{\textit{Seadronix} \\
{\tt\small \{yongjin, jin.park, ksana, hank05\}@seadronix.com}
}
}

\maketitle

\begin{abstract}
The maritime shipping industry is undergoing rapid evolution driven by advancements in computer vision artificial intelligence (AI). Consequently, research on AI-based object recognition models for maritime transportation is steadily growing, leveraging advancements in sensor technology and computing performance. However, object recognition in maritime environments faces challenges such as light reflection, interference, intense lighting, and various weather conditions. To address these challenges, high-performance deep learning algorithms tailored to maritime imagery and high-quality datasets specialized for maritime scenes are essential. Existing AI recognition models and datasets have limited suitability for composing autonomous navigation systems. Therefore, in this paper, we propose a Vertical and Detail Attention (VaDA) model for maritime object segmentation and a new model evaluation method, the Integrated Figure of Calculation Performance (IFCP), to verify its suitability for the system in real-time. Additionally, we introduce a benchmark maritime dataset, OASIs (Ocean AI Segmentation Initiatives) to standardize model performance evaluation across diverse maritime environments. OASIs dataset and details are available at our website: \href{https://www.navlue.com/dataset}{https://www.navlue.com/dataset}.
\end{abstract}


\IEEEpeerreviewmaketitle

\section{Introduction} \label{sec:introduction}
The maritime shipping industry has been transitioning into a smart maritime system recently driven by digitization, informatization, and now advancements in artificial intelligence (AI). Indeed, at the core of these changes is the advancement of computer vision technology tailored for autonomous navigation systems. These technological strides are not only revolutionizing smart ports but also reshaping the entire maritime industry. Research into AI-based object recognition models in maritime transportation has steadily increased over time. Advancements in sensor technology, computing power, and AI have accelerated the development of these models. Autonomous navigation systems based on computer vision must collect and analyze sensor data to make real-time decisions. In this context, the recognition performance and speed of AI models related to maritime object recognition are crucial aspects. These technological advancements play a crucial role in enhancing safety and efficiency in the maritime transportation industry. Accurate object recognition and tracking help minimize the risk of accidents and optimize overall operational performance.

\begin{figure}
    \centering
        \includegraphics[width=\columnwidth]{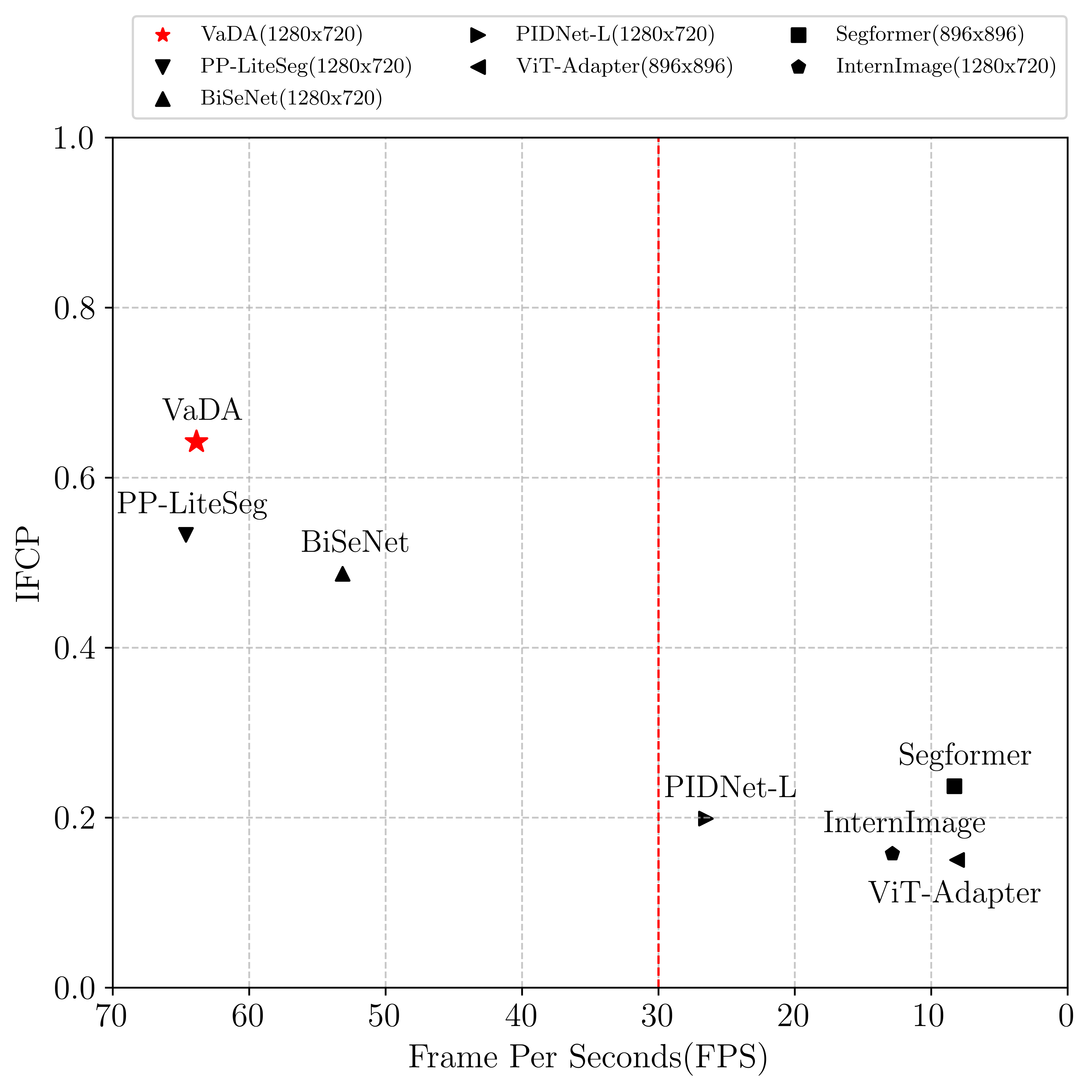}
    \caption{
    The comparison of Integrated Figure of Calculation Performance (IFCP) and Frame Per Seconds (FPS) on the Seadronix evaluation dataset, OASIs (Ocean AI Segmentation Initiatives). 
    Red marker represents our proposed Vertical and Detail Attention (VaDA). The test environment matches the inference environment.
    The experimental results demonstrate that VaDA achieves the best IFCP. }
\label{fig:speed_iou_fig}
\end{figure}

In marine environments, object recognition from images presents significant challenges due to factors such as light reflection, interference, intense illumination, and varying weather conditions. In addition, the object information may be distorted or noisy due to the image sensor or lens specifications. To address this issue, there is a need for research focused on developing high-performance deep-learning algorithms specifically tailored to the unique characteristics of maritime data. In addition, it is essential to construct high-quality datasets specialized in maritime-related images to effectively train and evaluate these algorithms.

The performance of deep learning models is strongly correlated with the quality and quantity of data as well as the complexity of the models. However, improvements in performance often come at the expense of increased model runtime. Lengthy runtime poses significant challenges for deploying models in real-world applications, where speed and efficiency are critical. Therefore, it is essential for the industry to prioritize the development of AI solutions that carefully balance performance and runtime considerations, ensuring their suitability for real-world environments.

Additionally, it is essential to construct high-quality datasets specialized in maritime-related images to effectively train and evaluate these algorithms. Thus, the following datasets have been proposed. The Bounding Box-based datasets such as Singapore Maritime~\cite{prasad2017video}, Seagull~\cite{ribeiro2017data}, SeaShips~\cite{shao2018seaships}, have been proposed for object detection and object tracking. Additionally, for semantic segmentation tasks, datasets like MaSTr1325~\cite{bovcon2019mastr}, MariShipSeg-HEU~\cite{zhang2020integrated}, Foggy ShipInsseg~\cite{sun2022irdclnet} have been proposed. Various research studies~\cite{hu2021pag, chen2021wodis, kim2022object, ribeiro2022real, chen2024efficientship} have been conducted utilizing these datasets. 

However, the proposed datasets are often either regional in scope or fail to represent various maritime conditions. This problem arises from the challenges associated with collecting data from various ships and ports, as well as the high costs involved in processing the collected data. In addition, existing proposed datasets often suffer from inaccurate labeling or are no longer maintained, making them unusable. Consequently, it has been established that there is still no suitable benchmark dataset for the maritime environment.

In this paper, we proposed Ocean AI Segmentation Initiatives dataset (OASIs) for evaluating segmentation performance in various marine environments. In addition, we propose a real-time segmentation model and suggest a novel evaluation metric for comparing it with other models. The new evaluation metric, Integrated Figure of Calculation Performance (IFCP), comprehensively assesses recognition accuracy, model parameters, GPU usage, and computational complexity. As illustrated in Figure 1, the proposed model, Vertical and Detail Attention (VaDA), demonstrates state-of-the-art performance in both Frame Per Second(FPS) and IFCP metrics.

\vspace{10pt}
\section{Related work} \label{sec:related_work}

\subsection{Maritime Datasets} \label{subsec:related_work_datasets}
Several datasets have been proposed to evaluate algorithms for detection and segmentation in marine environments. Fefilatyev et al.~\cite{fefilatyev2006horizon_buoy_dataset} proposed a dataset captured on the same day, including 10 sequences of open-view sea scenes. However, since this dataset is constructed to evaluate algorithms for detecting the horizon of the sea, the diversity of scenes is very limited and objects are not included.

As vision algorithms for object detection have developed a lot, several datasets for detecting objects/obstacles in the marine environment have also been proposed.
Kristan et al.~\cite{KristanCYB2015} introduced a dataset containing 12 different sequences captured by the USV, and Bovcon et al.~\cite{bovcon2018stereo} later extended it to a sea obstacle detection dataset containing large and small objects and a large number of horizontal lines using 28 stereo camera sequences synchronized with the Inertial Measurement Unit (IMU). Furthermore, this dataset contains many scenes of different weather and has a greater diversity of scenes.

Prasad et al.~\cite{smd_prasad2017video} proposed a multi-sensor acquisition dataset containing 51 RGB and 30 Near-Infrared (NIR) image sequences, but the scene diversity is also large because it is acquired under different weather conditions and on different days, but since it is a survival dataset, most of the sequences consisted of fixed terrestrial views and highly static scenes.

Ribeiro et al.~\cite{ribeiro2017data} proposed a multi-sensor acquisition dataset containing 51 RGB and 30 NIR image sequences. This dataset provides considerable scene diversity, as it was captured under various weather conditions and on different days. However, since it is a surveillance dataset, most of the sequences consist of fixed terrestrial views and highly static scenes.

Seaships~\cite{shao2018seaships} introduced 31,455 images with 1920$\times$1080 resolution including six types of ships as target objects. The data set was acquired at a set time through the inland waterway's monitoring video system. That's why all the scenes have a certain directional view.
The datasets currently available for marine environments focus on object detection models, with relatively fewer for segmentation. Recently, there have been proposals for benchmark datasets to evaluate the performance of segmentation models at sea. Bovcon et al.~\cite{bovcon2019mastr} proposed a dataset called ``MaSTr1325" which is a marine semantic segmentation training dataset specifically designed to advance obstacle detection techniques in small coastal USVs. It consists of 1,325 different images taken over a two-year period using USVs, the various realistic conditions encountered during coastal surveillance missions. While each image was carefully labeled pixel-wise and synchronized with an onboard sensor for semantic understanding, only three labels (sky, water, and obstacle/environment) were labeled by annotation.

Most recently, Bovcon et al.~\cite{bovcon2021mods} released an evaluation dataset that complements the previously published maritime segmentation datasets~\cite{KristanCYB2015, bovcon2018stereo, bovcon2019mastr}. This dataset consists of over 80k stereo images and has been recorded in multiple locations, including various obstacles. Measurements were made at various times and weather environments for about seven months.
The dataset is in two stages, with experts refining the per-pixel labeling following initial labeling tasks obtained from internet platforms. It is considered the most challenging benchmark dataset for marine environments because existing state-of-the-art models do not perform well on it.

\subsection{Semantic Segmentation}
It is important to consider the performance and inference time of deep learning models in developing solutions for real-world industrial applications; hence, we have focused on real-time and state-of-the-art performance segmentation models.\par
\vspace{3pt}
\subsubsection{Real-Time Segmentation} \label{subsec:related_work_real_time_segmentation}
It is natural in AI models that a trade-off occurs, in which the performance decreases as the model operates fast. For real-time operation of the model, there are very simple methods such as using a lightweight backbone or applying a limited-sized input image. 
However, this lightweight backbone of the classification model is not perfectly suited to semantic segmentation and also cannot extract clear features from the small objects. Thus, various studies have been proposed to operate segmentation models in real-time in a way that utilizes lightweight backbones or extracts sufficient features with low computing costs. 
Fan et al.~\cite{fan2021rethinking} proposed a backbone with a Short-Term Dense Concatenate (STDC) module that reduces composition costs and extracts rich features through scalable receptive fields and multi-scale information by reducing redundancy of two branch structures but failed to produce sufficient trade-off in terms of accountability and speed. 
BiSeNet~\cite{yu2018BiSeNet} presented a two-branch network, allowing large receptive fields of detail and contextual information that are important in semantic segmentation. However, it was still difficult to expect high performance of real-time models due to the structural latency and expensive computational costs in Two-Branch Network (TBN) architectures. Therefore, many studies~\cite{yu2018BiSeNet, yu2021BiSeNet} have been conducted in the direction of enriching deep features and reducing model computation costs while maintaining a TBN framework.
Motivated by the fact that the architecture of TBN is similar to that of PI controllers, PIDNet~\cite{xu2023pidnet}, a novel three-branch network, was proposed to solve the overshooting problem occurring in the existing TBNs~\cite{yu2018BiSeNet,yu2021BiSeNet,fan2021rethinking}.

\begin{figure}
    \centering
    \includegraphics[width=0.95\columnwidth]{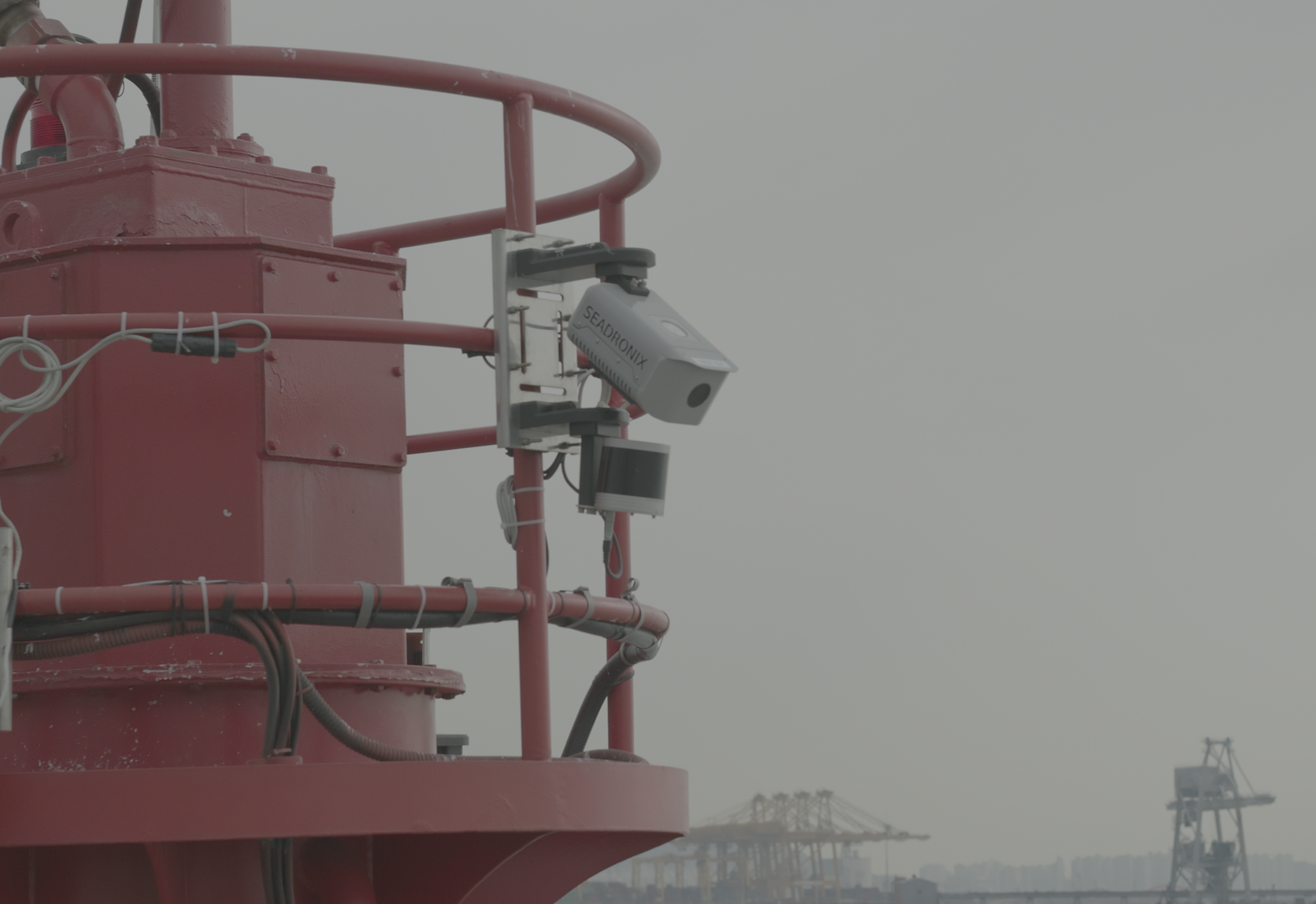}
    \caption{The sample image of our measuring Hardware equipment, \textbf{SxSM200N}.}
    \label{fig:CM_module}
\end{figure}

\begin{figure*}
    \centering
    \includegraphics[width=0.99\columnwidth]{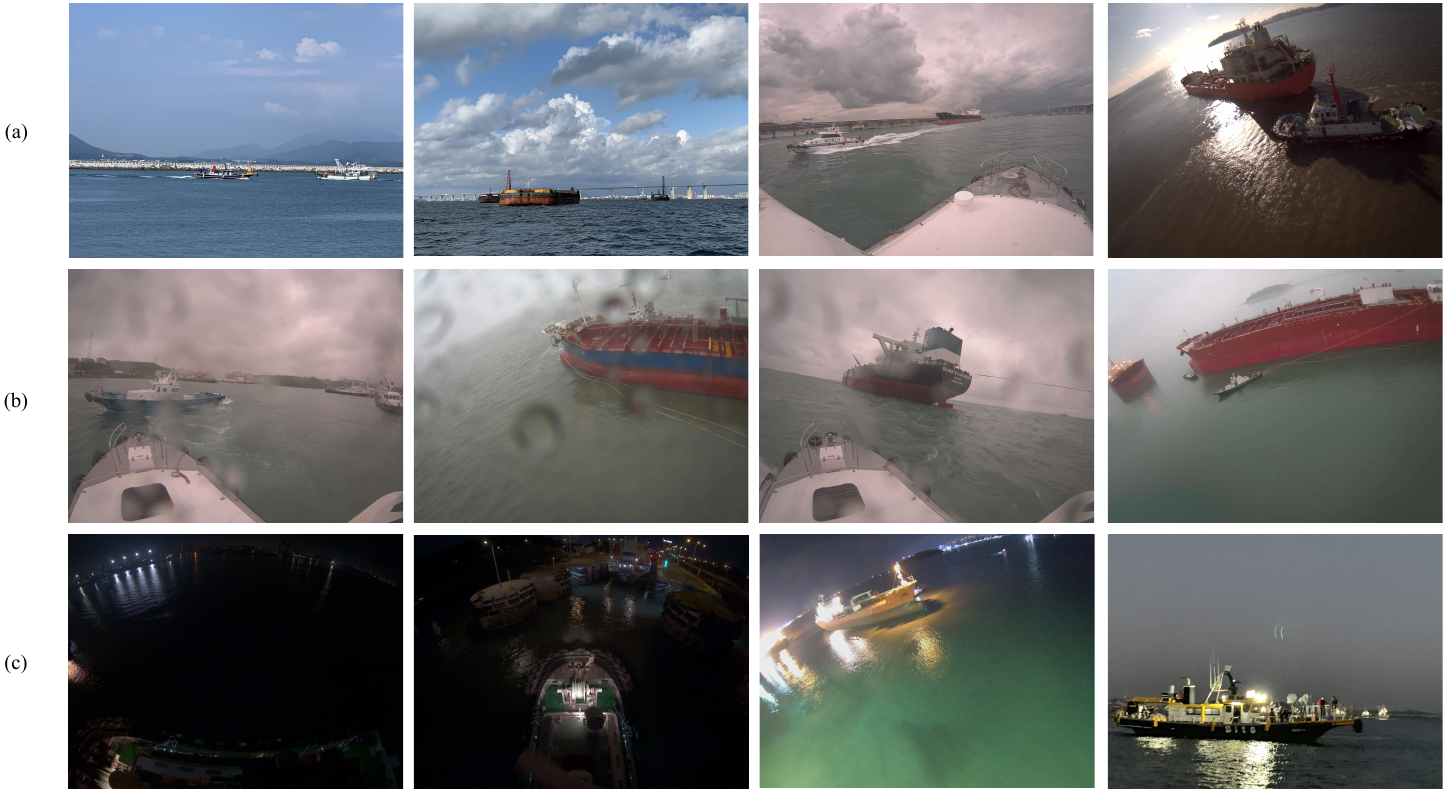}
    \caption{The sample images of OASIs (Ocean AI Segmentation Initiatives). (a) shows included daytime scenes (sunny, mild cloudy,back-lit) in OASIs Type-1. (b) includes abnormal weather scenes (rainy, foggy) in Type-2 and (c) includes night-time scenes (dark, dark w/ light source, early evening, and dawn) in Type-3 respectively. }
    \label{fig:dataset_sampleImgs}
\end{figure*}
\subsubsection{State-of-the-Art Segmentation}\label{subsec:related_work_sota_segmentation}
Driven by the great achievements of transformer~\cite{vaswani2017attention} in NLP, Vision Transformer (ViT)~\cite{dosovitskiy2020image} introduced a transformer architecture for image classification that processes input images as sequential patch tokens.
For semantic segmentation, SETR~\cite{zheng2021rethinking} adopts ViT~\cite{dosovitskiy2020image}  as a backbone for extracting features, achieving promising performance. Segmenter~\cite{strudel2021segmenter} proposes a transformer encoder-decoder architecture for semantic image segmentation. This approach relies on the backbone of ViT~\cite{dosovitskiy2020image} and introduces a mask decoder inspired by DETR~\cite{carion2020end}. PVT~\cite{wang2021pyramid} enriches features by constructing ViTs into pyramid structures, just as it extracts sufficient features through the connection of pyramid structures in CNN models. SegFormer~\cite{xie2021segformer} is designed to perform more simple and efficient segmentation tasks using hierarchically structured transformer encoders and lightweight decoders for multi-scale features. However, despite the high performance of these transformer-based methods~\cite{carion2020end, wang2021pyramid, xie2021segformer} the high cost makes them difficult to deploy in real-time applications.
Recently, InternImage~\cite{wang2023internimage} has designed convolutions in custom block-level architectures such as Transformers to design a CNN-based foundation model.  Using variations of flexible convolutions, called deformable convolutions (DCN)~\cite{dai2017deformable, zhu2019deformable}, it performs comparably to transformer-based models.
\vspace{10pt}
\section{OASIs: Ocean AI Segmentation Initiatives}\label{sec:proposed_dataset}
Our main goal was to build a comprehensive dataset for evaluating the performance of vision models operating in marine environments. The dataset mentioned above (Section~\ref{subsec:related_work_datasets}) is limited in diversity because it is difficult to collect data from various ships and ports, and it costs a lot to process the collected data. Therefore, we present OASIs, a more realistic and comparable important dataset of marine environments.
First, we explain our data collection system and method (Section~\ref{sec:dataset_acquisition}).
We next describe how the dataset images are processed (Section~\ref{sec:dataset_annotation}) and how it is selected and 
configured (Section~\ref{sec:dataset_configuration}).

\begin{figure}
    \centering
    \includegraphics[width=0.99\columnwidth]{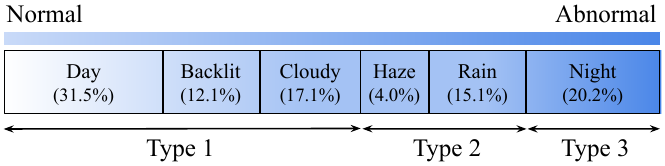}
    \caption{The scene distribution of maritime weather conditions in our evaluation dataset, OASIs. Normal weather conditions are on the left side of the distribution bar and Abnormal weather conditions are on the right. ``\textbf{Night}" contains night time condition scenes that are difficult for the RGB sensor of the camera to obtain information.}
    \label{fig:dataset_config_type_weather}
\end{figure}

\begin{table}[]
    \begin{tabular}{
    >{\columncolor[HTML]{FFFFFF}}c c}
    Class                                                    & \cellcolor[HTML]{FFFFFF}Grayscale \\ \hline
    \multicolumn{1}{c|}{\cellcolor[HTML]{FFFFFF}Others}      & 0                                 \\
    \multicolumn{1}{c|}{\cellcolor[HTML]{FFFFFF}Sea}         & 50                                \\
    \multicolumn{1}{c|}{\cellcolor[HTML]{FFFFFF}Land}        & 100                               \\
    \multicolumn{1}{c|}{\cellcolor[HTML]{FFFFFF}Sea Objects} & 150                              
    \end{tabular}
    \caption{The table of annotated colors in each label.}
    \label{table:color_label}
\end{table}

\subsection{Dataset Acquisition}\label{sec:dataset_acquisition}

In this paper, we constructed a dataset by acquiring images related to the marine environment using the sensor module ``SxSM200N'' (Figure~\ref{fig:CM_module}) from 2017 to 2023. This allowed us to gather images from various berths and ships, enabling us to construct an OASIs that encompasses a wide range of environmental variables. Unlike existing datasets that capture the sea at specific locations and times, resulting in limited scenes and styles, our proposed dataset encompasses a broader range of environmental conditions. The collection locations are major ports and waterways, including Ulsan and Busan in South Korea.

\subsection{Dataset Annotation}\label{sec:dataset_annotation}

\begin{figure}
    \centering
    \includegraphics[width=0.99\columnwidth]{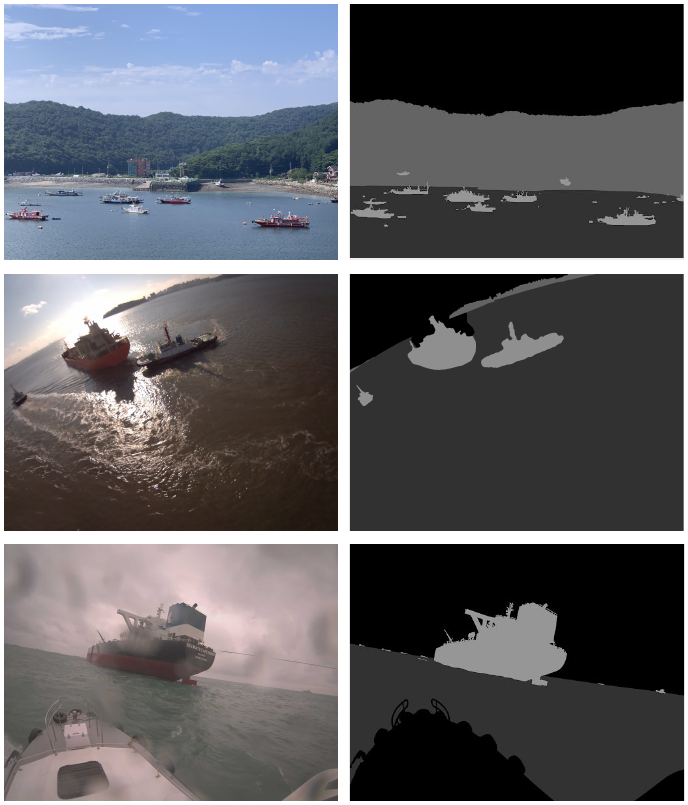}
    \caption{The sample images and annotated labels pairs. (Left) shows an original image of OASIs. (Right) shows an annotated label for each input image. }
    \label{fig:sample_annotation}
\end{figure}

The images in the OASIs varies in resolution, ranging from 1280$\times$720 to 4032$\times$3024 pixels. The dataset, used for both training and evaluation, is labeled by experts. The images provided to the experts are selected from real-time captures that meet specific quality standards. This labeling process is consistently performed at the pixel level according to internally established guidelines. Semantic segmentation labels of OASIs are annotated to provide pixel-level classification, where each pixel in an image is assigned a color label corresponding to a particular class. Figure~\ref{fig:sample_annotation} shows a sample labeled image created following these guidelines. Each class is described in grayscale, as detailed in Table~\ref{table:color_label}.\par


\subsection{Dataset Configuration}\label{sec:dataset_configuration}

The marine environment exhibits unique characteristics, with rapidly and dramatically changing weather conditions that highlight its distinct features. However, existing benchmark datasets do not contain images of different weather environments and times, making it difficult to grasp the general performance of the models. Large-scale detection and segmentation learning datasets such as COCO~\cite{lin2014coco} and PASCAL VOC~\cite{pascal-voc-2012-xkk13_dataset} also lack images depicting various weather conditions at sea, even though they include a small number of images featuring marine environments and different weather scenarios. Several recent studies~\cite{song2023extreme, fischer2015anthropogenic} in the marine field have shown that the intensity and frequency of various extreme weather events have increased in the marine environment. Chen et al.~\cite{chen2024weather} found that mAP values of DETR~\cite{carion2020end} appear with significant performance degradation of 93\% and 78\%, respectively, when synthesized rain-noise and haze-noise are added to existing data.
In this paper, we propose a marine environment dataset OASIs, which is categorized into three types: daytime, weather conditions, and nighttime. Through these segmented datasets, the performance of AI models can be evaluated in each environment. Figure~\ref{fig:dataset_config_type_weather} shows the ratio of adverse weather conditions in OASIs. OASIs was divided into three types as shown in Table~\ref{table:eval_dataset_config}.
The detailed characteristics of each type are described below.

\vspace{3pt}

\begin{figure}
    \centering
    \includegraphics[width=0.95\columnwidth]{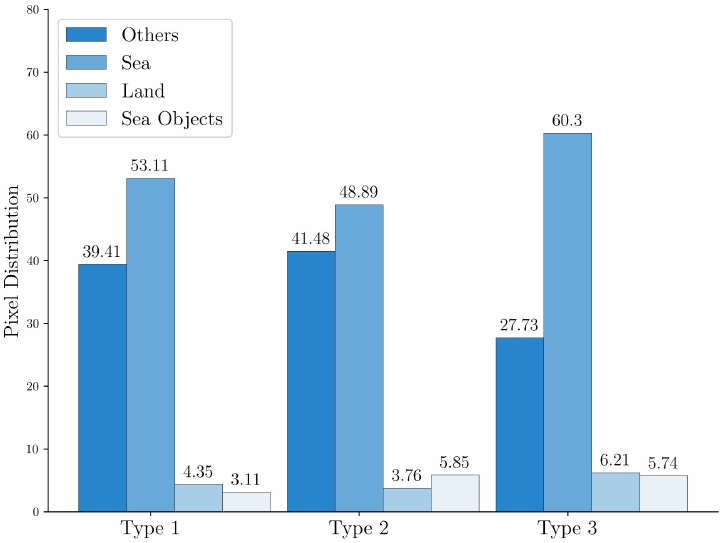}
    \caption{The average pixel distribution of images in each type of OASIs. The \textit{label \textbf{``Others"}} includes all the abnormal maritime weather environments.}
    \label{fig:bars_dataset}
\end{figure}
\begin{table}[]
    \begin{tabular}{
    >{\columncolor[HTML]{FFFFFF}}c c}
    \multicolumn{1}{l|}{\cellcolor[HTML]{FFFFFF}Type-1}      & Normal Condition, Backlit, Cloudy        \\ \hline
    \multicolumn{1}{l|}{\cellcolor[HTML]{FFFFFF}Type-2}         & Rain, Haze                                \\ \hline
    \multicolumn{1}{l|}{\cellcolor[HTML]{FFFFFF}Type-3}        & Night, Night light, Early night and Dawn      \\          
    \end{tabular}
    \caption{The table of criteria to divide datasets by weather and environment.}
    \label{table:eval_dataset_config}
\end{table}
\vspace{3pt}
\subsubsection{Type-1: Day-Time Environment}
The most basic environment images collected during the daytime are included in the data Type 1. It contains data from general weather environments including normal lights, backlit, and several mild cloudy situations. Therefore, long-distance objects can be seen well, also it is an environment that can distinguish objects well and the boundary between the sea and the sky can be clearly distinguished. However, it also includes some scenes where the characteristics of a specific object are lost due to backlit, or where it is difficult to distinguish long-distance objects due to cloudy weather. \par
\vspace{3pt}
\subsubsection{Type-2: Adverse Weather Environment}
In the case of Type-2, images collected in poor-weather environments such as rain and fog are included. As mentioned above, Collecting images in a marine environment are exposed to various weather environments (rain, sea fog, snow, and others). Type 2 includes foggy and rainy data. In a haze situation, long-distance objects are faint or only partially visible. In rainy conditions, raindrops cover the lens, and these physical constraints show features of unclear boundaries between the sky, the ocean, the ground, and objects.\par
\vspace{3pt}
\subsubsection{Type-3: Night-Time Environment}
To effectively utilize the vision model, it must perform well not only during the day but also at night when lighting conditions are insufficient. Type 3 includes images captured during the evening hours at ports and on ships. The evening maritime environment, with its significant lack of visible light, poses a substantial challenge for RGB sensors to capture information effectively. In these images, features for object identification are often barely visible, and sensor pixel saturation frequently occurs due to strong light. Additionally, the boundaries between the sea, land, and sky are often indistinct.\par
\vspace{3pt}
The three divided datasets vary in scene environments, including weather, lighting, and time. However, an analysis of the labeled pixel distribution reveals a similar pattern across all datasets. The label ``Sea'' is the most prevalent, followed by ``Others''. The label ``Land'' and ``Sea Objects'' each constitute less than 7$\%$ of the total image pixels. This indicates that our evaluation dataset, OASIs is well balanced in terms of annotated labels, as illustrated in Figure~\ref{fig:bars_dataset}.

\vspace{10pt}
\section{VaDA: Vertical and Detail Attention} \label{sec:proposed_method}\par
In this paper, we proposed a model for improving image recognition performance in marine environments, focusing on the following areas:
\subsection{Feature Extraction Backbone}\label{sec:vada_featurebackbone}
The backbone of the proposed model is designed to extract robust features from various conditions present in marine environments. It has been proven effective in extracting features from factors such as strong lighting, sunlight-induced sea surface reflection, and interference. This allows the model to significantly enhance the accuracy of object recognition in marine environments. Additionally, the model is designed with fewer parameters to ensure fast inference speed.
\subsection{Attention Module}\label{sec:vada_attention_module}
Since data collected through installed cameras typically exist objects horizontally, an attention module operating vertically is added to the proposed model to distinguish horizontally existing objects. Particularly, this attention module greatly enhances the recognition performance of crucial elements, such as horizontal lines, in identifying maritime situations.
\subsection{Loss Function}\label{sec:vada_lossfunction}
The model was trained using detail loss to ensure stable recognition of object edges even in images with diverse weather conditions and camera noise. This approach led to improvements in performance both in recognizing the edge portions of small objects under perspective and in capturing the detailed shapes of objects.

Through the proposed methods, this study aims to contribute to the enhancement of image recognition performance in marine environments.

\label{sec:semantic_segmentation_VaDA}
\begin{table*}[htb]
\begin{tabular}{llrrrrll}
\hline
\multicolumn{1}{c}{Model} & \multicolumn{1}{c}{Batch} & \multicolumn{1}{c}{1} & \multicolumn{1}{c}{2} & \multicolumn{1}{c}{4} & \multicolumn{1}{c}{6} & \multicolumn{1}{c}{8} & \multicolumn{1}{c}{10}\\ \hline
\multirow{3}{*}{\begin{tabular}[c]{@{}l@{}}VaDA\\ (1280x720)\end{tabular}}& FLOPS{[}GB{]} & 72.00& 144.00 & 288.00 & 432.00 & \multicolumn{1}{r}{576.00} & \multicolumn{1}{r}{720.00} \\
&GPU mem usage{[}GB{]}& 3.58 & 6.98 & 13.80& 20.78& \multicolumn{1}{r}{27.71}& \multicolumn{1}{r}{34.62} \\
& GPU time{[}ms{]}& 15.66& 28.03& 52.60& 76.57& \multicolumn{1}{r}{101.31} & \multicolumn{1}{r}{125.73} \\ \hline
\multirow{3}{*}{\begin{tabular}[c]{@{}l@{}}PP-LiteSeg\\ (1280x720)\end{tabular}}& FLOPS{[}GB{]} & 76.95& 153.89 & 307.79 & 461.68 & & \\
& GPU mem usage{[}GB{]} & 4.16 & 8.11 & 16.02& 24.05& & \\
& GPU time{[}ms{]}& 15.47& 28.32& 54.01& 78.87& & \\ \hline
\multirow{3}{*}{\begin{tabular}[c]{@{}l@{}}BiSeNet\\ (1280x720)\end{tabular}} & FLOPS{[}GB{]} & 89.56& 179.12 & 358.23 & 537.35 & & \\
& GPU mem usage{[}GB{]} & 3.50 & 6.80 & 13.41& 20.11& & \\
& GPU time{[}ms{]}& 18.82& 36.57& 69.62& 98.05& & \\ \hline
\multirow{3}{*}{\begin{tabular}[c]{@{}l@{}}PIDNet-L\\ (1280x720)\end{tabular}}& FLOPS{[}GB{]} & 253.19 & 506.37 & 1012.74 & & & \\
& GPU mem usage{[}GB{]} & 6.85 & 13.26 & 26.10 & & & \\
& GPU time{[}ms{]}& 37.71& 69.71& 132.86 &  & & \\ \hline
\multirow{3}{*}{\begin{tabular}[c]{@{}l@{}}ViT-Adapter\\ (896x896)\end{tabular}}& FLOPS{[}GB{]} & 208.93 & 729.30 & 1458.59& \multicolumn{1}{l}{}& & \\
& GPU mem usage{[}GB{]} & 22.78& 30.34& 60.49& \multicolumn{1}{l}{}& & \\
& GPU time{[}ms{]}& 123.35 & 166.80 & 334.71 & \multicolumn{1}{l}{}& & \\ \hline
\multirow{3}{*}{\begin{tabular}[c]{@{}l@{}}Segformer-B1\\ (896x896)\end{tabular}} & FLOPS{[}GB{]} & 350.84 & \multicolumn{1}{l}{}& \multicolumn{1}{l}{}& \multicolumn{1}{l}{}& & \\
& GPU mem usage{[}GB{]} & 15.17& \multicolumn{1}{l}{}& \multicolumn{1}{l}{}& \multicolumn{1}{l}{}& & \\
& GPU time{[}ms{]}& 120.49 & \multicolumn{1}{l}{}& \multicolumn{1}{l}{}& \multicolumn{1}{l}{}& & \\ \hline
\multirow{3}{*}{\begin{tabular}[c]{@{}l@{}}Internimage\\ (1280x720)\end{tabular}} & FLOPS{[}GB{]} & 364.65 & \multicolumn{1}{l}{}& \multicolumn{1}{l}{}& \multicolumn{1}{l}{}& & \\
& GPU mem usage{[}GB{]} & 14.89& \multicolumn{1}{l}{}& \multicolumn{1}{l}{}& \multicolumn{1}{l}{}& & \\
& GPU time{[}ms{]}& 77.85& \multicolumn{1}{l}{}& \multicolumn{1}{l}{}& \multicolumn{1}{l}{}& & \\ \hline
\end{tabular}
\caption{Analysis of FLOPS, GPU usage, and GPU time metrics based on batch size in inference environment. VaDA can utilize the largest batch size for training given the same computational resources as other models.} 
\label{table:gpu_performance_comparison}
\end{table*}

\section{Experiments}\label{sec:Experiments}
In this section, we train the proposed model and state-of-the-art models using the training dataset provided by Seadronix Corp. We then compare and analyze their performance evaluation on the proposed Seadronix evaluation dataset, OASIs. The key metrics used to evaluate the models are as follows: \
\subsubsection{Intersection over Union (IoU)}
An essential metric for evaluating accuracy in image segmentation. It calculates the ratio of the intersection area between the predicted segmentation mask and the ground truth mask to the union area of the predicted and ground truth areas.\
\subsubsection{Integrated Figure of Calculation Performance (IFCP)}
There are various metrics used to measure the performance of AI models. However, there isn't a single comprehensive performance metric that considers all of these metrics. In existing research, models' performance has been evaluated by comparing metrics such as CPU time, GPU time, mIOU, Multiply-Adds, Params, and Latency in a single table\cite{howard2019searching}, \cite{cai2018proxylessnas}. However, this makes it difficult to judge the model's performance at a glance.
To address this issue, we introduced a comprehensive metric named Integrated Figure of Calculation Performance (IFCP) to evaluate the overall performance of the model. IFCP is a metric that considers IoU, FLOPS[GB], GPU usage[GB], and parameters of models (Params[MB]). The calculation method of IFCP is as follows: The IoU value and other parameters are divided individually. Then, the harmonic mean (H) of these values is calculated. The formula for the harmonic mean is as follows:
\begin{align*}
    H=\frac{n}{\frac{1}{x_0}+\frac{1}{x_1} \cdot \cdot \cdot \frac{1}{x_n}}=\frac{n}{\sum_{i=0}^{n}\frac{1}{x_i}}
\end{align*}
Through this process, the difference in measurement units of each parameter is adjusted, ensuring that each parameter is equally weighted. This computation enables fair contribution from all parameters and ensures an accurate evaluation of overall performance.
It is defined by the following equation:
\begin{align*}
	IFCP & =\frac{3}{\frac{1}{\frac{IoU}{FLOPS}}+\frac{1}{\frac{IoU}{GPU\;\;usage}}+\frac{1}{\frac{IoU}{Params}}}\\\\
    	   & =\frac{3 \times IoU}{FLOPS+GPU\;\;usage+Params} 
\end{align*}

\begin{itemize}
\item IoU: Intersection over Union
\end{itemize}
\begin{itemize}
\item FLOPS: The total number of floating-point operations performed by the model in a single forward pass.
\end{itemize}
\begin{itemize}
\item GPU usage: The total amount of GPU memory consumed by operations or function calls.
\end{itemize}
\begin{itemize}
\item Params: The number of parameters in the model.
\end{itemize}

\vspace{3pt}
The baseline benchmarks for IFCP are established as follows: a mIoU (mean Intersection over Union) of 95\% to represent human-level perception, the maximum GPU memory available for edge devices is typically 8GB, a model size of 10MB and 50GB FLOPS for deployment on edge devices. IFCP comprehensively evaluates the performance of the model considering accuracy, GPU memory usage, computational complexity, and model size.  As a result, this metric facilitates a holistic assessment of recognition performance and speed. also, it is possible to confirm whether the designed model can operate in real-time and perform accurate recognition on edge devices deployed in the marine environment. 

\subsection{Implementation Details} \label{sec:implementation_details}
\subsubsection{Training}In this paper, pretraining was applied to the models being compared according to the methods proposed in their respective papers. For the training protocols, efforts were made to maintain common settings across all models. The Rectified Adam (RAdam\cite{liu2019variance}) algorithm was chosen as the optimizer. The training was conducted for 100 epochs with an initial learning rate of \num{1e-3}. Augmentation strategies were kept consistent as common training parameters across all models. For CNN-based models such as VaDA, PP-LiteSeg\cite{fan2021rethinking}, PIDNet\cite{xu2023pidnet}, BiSeNet\cite{yu2018BiSeNet}, and InternImage\cite{wang2023internimage}, the training image size was set to 1280$\times$720. For self-attention-based models, the training image size was set to 896$\times$896.\ 

\begin{figure}
    \centering
        \includegraphics[width=\columnwidth]{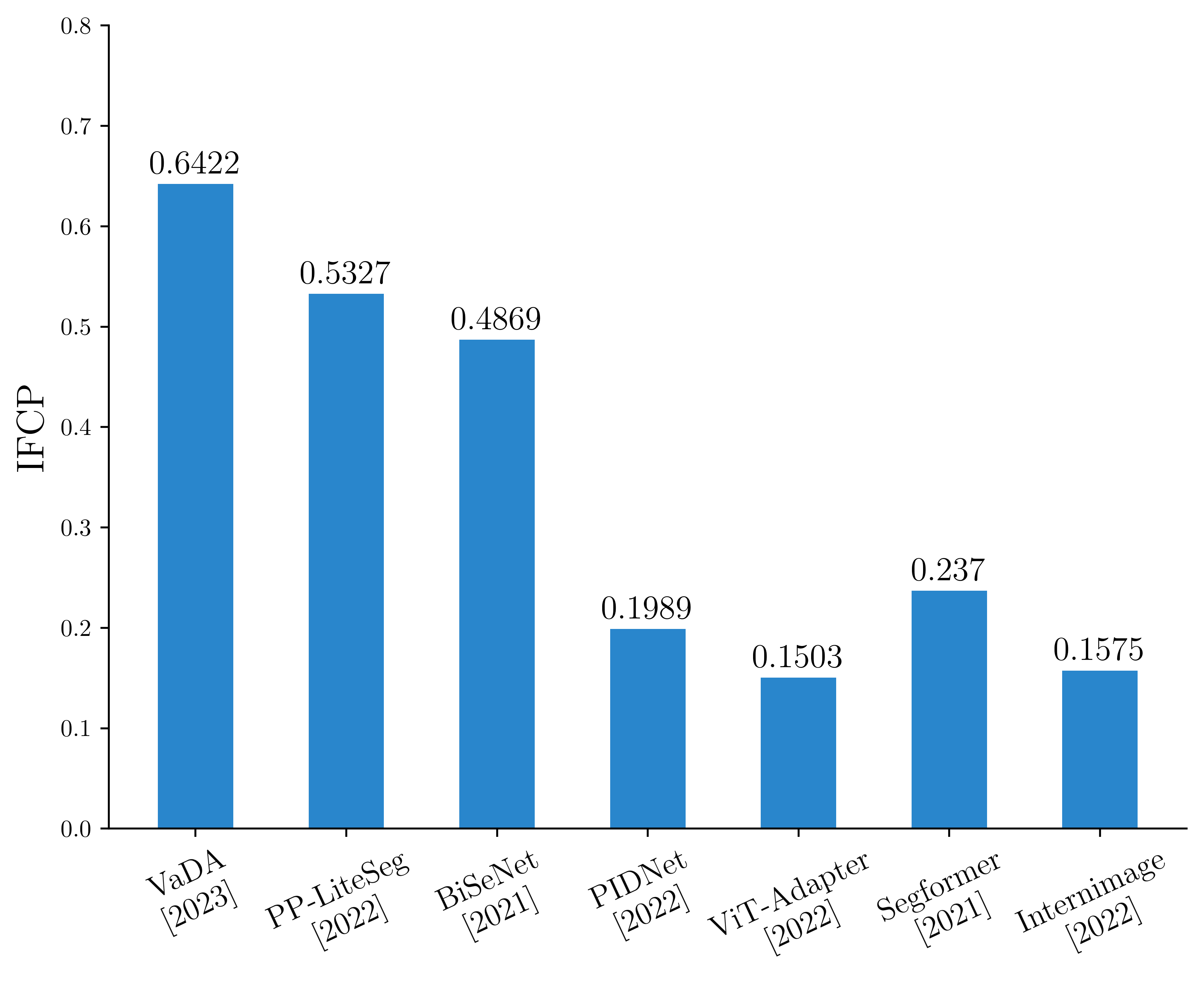}
    \caption{The IFCP value comparison with \textbf{real-time} and \textbf{state-of-the-art} models on our proposed dataset, OASIs. Higher values indicate superior model performance. }
\label{fig:ifcp_fig}
\end{figure}

\vspace{3pt}
\subsubsection{Inference}
Inference speed was measured on a platform of NVIDIA TU104 based GPU, PyTorch 1.8 under Ubuntu. 
Batch normalization was integrated into each convolution layer, and the batch size was set to 1 for pair comparison. In the inference environment, the Frame Per Sec(FPS) of each model is as follows: [VaDA : 63.87, PP-LiteSeg : 64.64, BiSeNet : 53.15, PIDNet-L : 26.52, Internimage : 12.85, Segformer-B1 : 8.30, ViT-Adapter : 8.11]


\subsection{Comparison of Model Computational Complexity}
\label{sec:model_complexity}
Table~\ref{table:gpu_performance_comparison} provides a comparative analysis of metrics, including FLOPS, GPU usage, and GPU time, for models utilized in the inference environment, categorized by batch size:
\begin{itemize}
\item FLOPS: The total amount of FLOPS.
\end{itemize}
\begin{itemize}
\item GPU mem usage: The total amount of GPU memory consumed by operations or function calls.\
\end{itemize}
\begin{itemize}
\item GPU time: The total GPU time spent in the operation or function call and its sub-calls.\
\end{itemize}

By analyzing the table, we can understand how the model's operational characteristics and memory usage vary with batch size. 
In addition, we can determine the maximum batch size by examining the empty cells in the table. In the case of VaDA, it is evident that it can operate with up to 10 batches. This demonstrates that VaDA exhibits maximum operational efficiency in constrained environments.

\subsection{Performance on evaluation OASIs}
\label{sec:performance_evaluation_eval_dataset}

IFCP is a metric that demonstrates the overall performance of the model. Through this metric, a comprehensive assessment of the applicability of AI models on edge devices deployed in maritime environments can be made. Since IFCP evaluates not only recognition performance but also the overall metrics of the model, it's notable that transformer models and large models, which are unsuitable for edge devices, exhibit low IFCP values. In contrast, the proposed VaDA model exhibited the highest performance, with a value of 0.6422, as depicted in Figure~\ref{fig:speed_iou_fig}. This suggests that when deployed on edge devices in maritime environments, the VaDA model is expected to demonstrate excellent performance.

\begin{table}[]
\begin{tabular}{lrr} \hline 
     & \multicolumn{1}{l}{Mean IoU} & \multicolumn{1}{l}{FPS} \\ \hline
VaDA & \textbf{0.7933}  & 63.87\\ \hline
PP-LiteSeg & 0.6962     & \textbf{64.64}   \\ \hline
BiSeNet    & 0.6436     & 53.15\\ \hline
\end{tabular}
    \caption{Comparative analysis of the speed and accuracy of real-time models across the entire OASIs dataset (Type-1, 2, 3).}
    \label{table:performance_comparison_mIoU_FPS}
\end{table}

\begin{table*}[]
\begin{tabular}{clrrrrr}
\hline
\multicolumn{1}{l}{Type} & Model& \multicolumn{1}{l}{Others} & \multicolumn{1}{l}{Sea} & \multicolumn{1}{l}{Land} & \multicolumn{1}{l}{Object} & \multicolumn{1}{l}{Mean IoU} \\ \hline
\multirow{3}{*}{1} & VaDA & 0.9394   & 0.9850& 0.7863 & 0.6248   & \textbf{0.8339}     \\
 & PP-LiteSeg & 0.9171   & 0.9616& 0.5565 & 0.4147   & 0.7125     \\
 & BiSeNet     & 0.8595   & 0.9196& 0.3915 & 0.2703   & 0.6102     \\ \hline
\multirow{3}{*}{2} & VaDA & 0.9874   & 0.9903& 0.8332 & 0.5696   & \textbf{0.8451}     \\
 & PP-LiteSeg & 0.9252   & 0.8963& 0.5151 & 0.4168   & 0.6883     \\
 & BiSeNet    & 0.8807   & 0.8653& 0.3994 & 0.3185   & 0.6160     \\ \hline
\multirow{3}{*}{3} & VaDA & 0.9784   & 0.9749& 0.8058 & 0.6227   & \textbf{0.8455}     \\
 & PP-LiteSeg & 0.6776   & 0.7722& 0.3094 & 0.2913   & 0.5126     \\
 & BiSeNet    & 0.8973   & 0.9244& 0.5189 & 0.4782   & 0.7047     \\ \hline
\end{tabular}
    \caption{The performance comparison with \textbf{real-time} ($\ge$30FPS) \textbf{state-of-the-art} models on OASIs.}
    \label{table:performance_comparison_real_time}
\end{table*}

\begin{table*}[]\begin{tabular}{clrrrrr}
\hline
\multicolumn{1}{l}{Type} & Model & \multicolumn{1}{l}{Others} & \multicolumn{1}{l}{Sea} & \multicolumn{1}{l}{Land} & \multicolumn{1}{l}{Object} & \multicolumn{1}{l}{Mean IoU} \\ \hline
\multirow{5}{*}{1} & VaDA  & 0.9394   & 0.9850& 0.7863 & 0.6248   & 0.8339     \\
 & ViT-Adapter & 0.9440   & 0.9859& 0.7665 & 0.7202   & 0.8542     \\
 & Internimage & 0.9582   & 0.9866& 0.7968 & 0.7393   & \textbf{0.8703}     \\
 & Segformer   & 0.9383   & 0.9781& 0.7106 & 0.6720   & 0.8248     \\
 & PIDNet-L    & 0.9298   & 0.9839& 0.7576 & 0.7003   & 0.8429     \\ \hline
\multirow{5}{*}{2} & VaDA  & 0.9874   & 0.9903& 0.8332 & 0.5696   & \textbf{0.8451}     \\
 & ViT-Adapter & 0.9793   & 0.9751& 0.7263 & 0.6579   & 0.8346     \\
 & Internimage & 0.9809   & 0.9761& 0.7295 & 0.6461   & 0.8332     \\
 & Segformer   & 0.9743   & 0.9773& 0.6841 & 0.6019   & 0.8094     \\
 & PIDNet-L    & 0.9780   & 0.9716& 0.6796 & 0.6317   & 0.8152     \\ \hline
\multirow{5}{*}{3} & VaDA  & 0.9784   & 0.9749& 0.8058 & 0.6227   & 0.8455     \\
 & ViT-Adapter & 0.9702   & 0.9738& 0.7753 & 0.7209   & 0.8601     \\
 & Internimage & 0.9733   & 0.9774& 0.7874 & 0.7444   & \textbf{0.8706}     \\
 & Segformer   & 0.9703   & 0.9770& 0.7909 & 0.7278   & 0.8665     \\
 & PIDNet-L    & 0.9693   & 0.9763& 0.7776 & 0.7240   & 0.8618     \\ \hline
\end{tabular}
    \caption{The performance comparison with \textbf{state-of-the-art} models on OASIs.}
    \label{table:performance_comparison_sota}
\end{table*}

The performance of the proposed VaDA model was evaluated through a comparison of real-time semantic segmentation models. Table~\ref{table:performance_comparison_mIoU_FPS} presents the IoU and FPS in the evaluation dataset, OASIs, of real-time models. The mIoU performance of the proposed VaDA is the best at 0.7993. In terms of FPS, it is 0.772 slower than PP-LiteSeg, but this difference is negligible. Table~\ref{table:performance_comparison_real_time} presents the mean IoU performance of each model for various types of OASIs. The VaDA model demonstrates the highest IoU values across all data types and maintains superiority even when compared on a per-class basis.  Furthermore, Table~\ref{table:performance_comparison_sota} illustrates the comparison results of the VaDA model with state-of-the-art models. In this comparison, the VaDA model exhibits the highest IoU values for Type-2 data. This finding indicates that the VaDA model delivers superior performance even in adverse weather conditions. Thus, it can be concluded that utilizing the VaDA model is advantageous, especially in maritime scenarios with rapidly changing weather environments.

\begin{figure*}
    \centering
    \includegraphics[width=0.99\columnwidth]{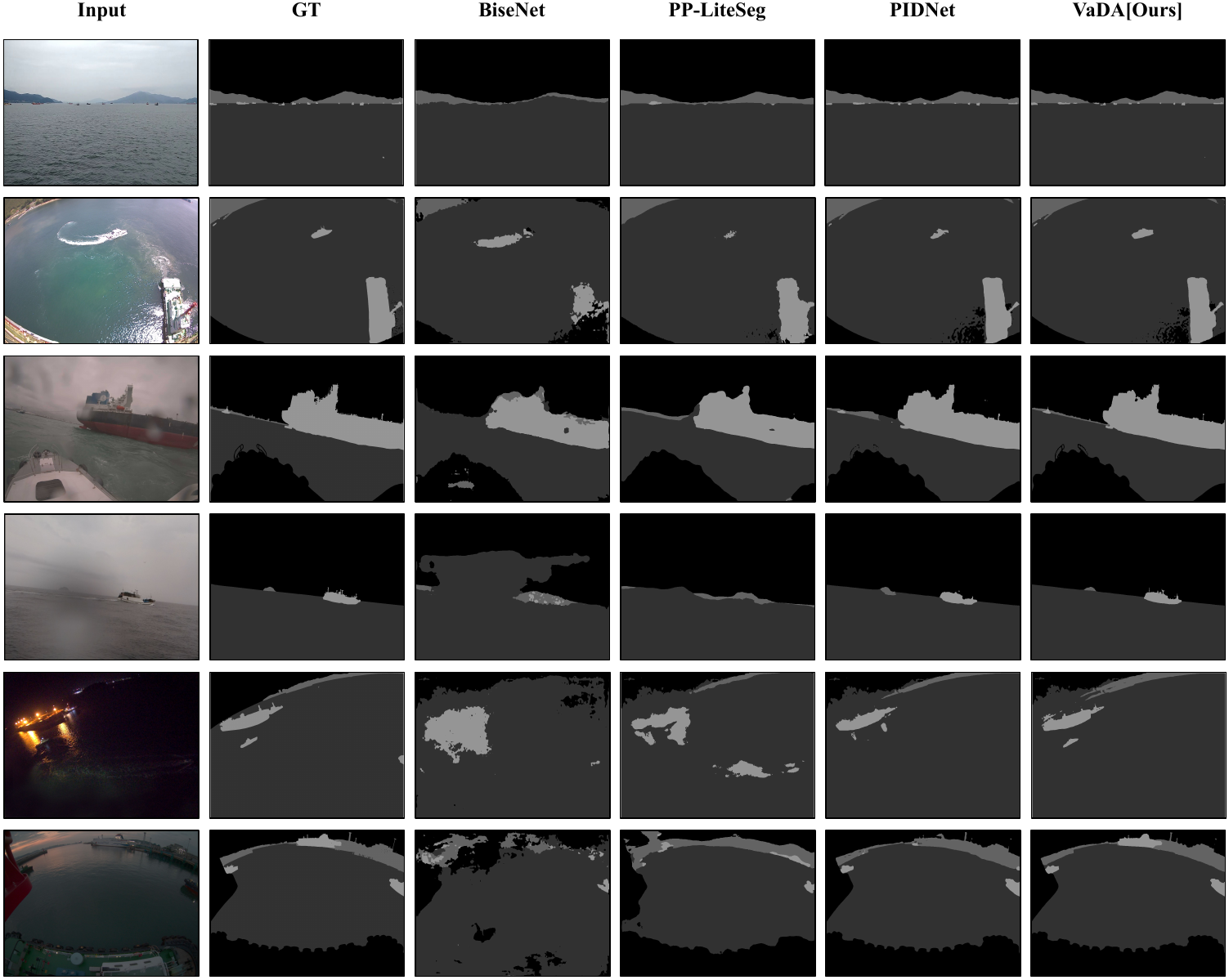}
    \caption{The comparison of qualitative results for \textbf{real-time} models on OASIs Type-1, 2 and 3. Compared to other models, \textbf{VaDA} consistently represents object boundaries across various environments and demonstrates a strong understanding of scene context.}
    \label{fig:dataset_segresults_fig1}
\end{figure*}

\begin{figure*}
    \centering
    \includegraphics[width=0.99\columnwidth]{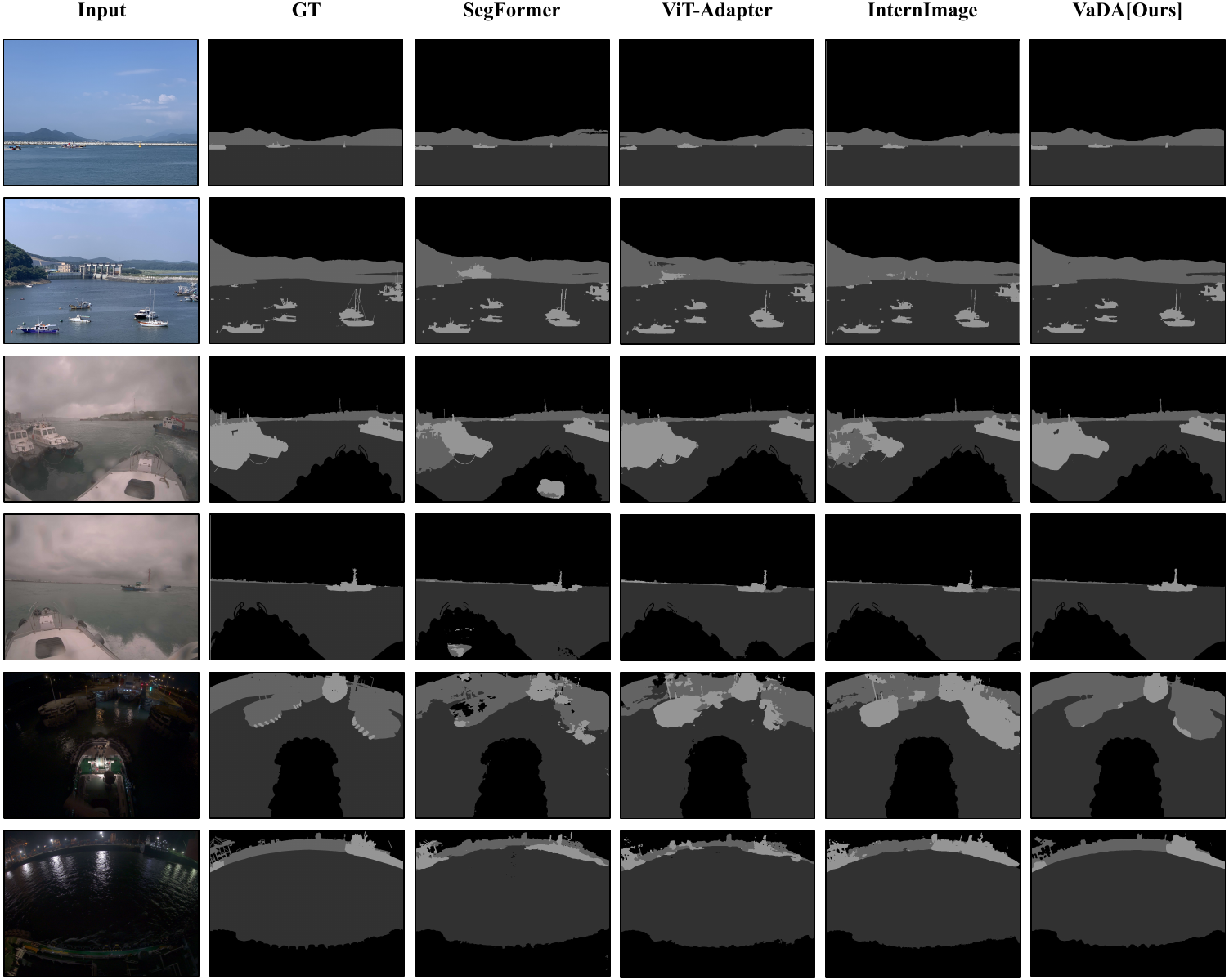}
    \caption{The comparison of qualitative results for \textbf{state of the art} on OASIs Type-1, 2 and 3. Compared to other models, \textbf{VaDA} consistently represents object boundaries across various environments and demonstrates a strong understanding of scene context. }
    \label{fig:dataset_segresults_fig2}
\end{figure*}

Figure~\ref{fig:dataset_segresults_fig1} displays the inference images of the real-time model, while Figure~\ref{fig:dataset_segresults_fig2} compares the inference results of the state-of-the-art model with the proposed VaDA model. In each figure, the leftmost column shows the original image and the ground truth (GT) image, while the rightmost column shows the inference results of the proposed VaDA model.

In Figure~\ref{fig:dataset_segresults_fig1}, it can be observed that the existing real-time model fails to properly distinguish the features of objects and does not accurately differentiate between objects. Similarly, in Figure~\ref{fig:dataset_segresults_fig2}, the performance of the VaDA model in inference is comparable to or even better than that of the state-of-the-art models, demonstrating promising recognition results.


\vspace{10pt}
\section{Conclusion}
\label{sec:conclusion}
Existing large-scale benchmark datasets such as COCO [31], PASCAL VOC [32], ADE20K [39], Cityscapes [40], and KITTI [41] are suitable for evaluating segmentation models in autonomous vehicle applications. However, they lack sufficient representation of maritime environments, making them inadequate for evaluating segmentation models specifically designed for these settings. To address this gap, we annotated a dataset using images gathered from ``SxSM200N'' deployed in ports and aboard ships, following internal guidelines. The OASIs is better suited for evaluating segmentation models in maritime environments and will significantly facilitate the comprehensive evaluation of segmentation models for recognizing maritime objects under diverse conditions.
Moreover, the VaDA model demonstrated remarkable recognition performance under varied weather conditions and proved to be highly suitable for deployment on edge devices. Utilizing the OASIs dataset, VaDA exhibited superior segmentation accuracy compared to existing state-of-the-art real-time models. Additionally, it outperformed others in the newly proposed model evaluation metrics.

These research findings highlight the potential of the VaDA model as an efficient AI solution for real-world applications. Additionally, these results represent pioneering work in the design and evaluation of computer vision AI systems specifically tailored for maritime environments.

The OASIs dataset and details are available at our website: \href{https://www.navlue.com/dataset}{https://www.navlue.com/dataset}. While the OASIs benchmark dataset primarily serves as an evaluation resource for marine environments, we plan to release additional datasets obtained from various sensors installed on ports and vessels where our products are deployed. Furthermore, we are researching and developing a multimodal model that complements the deep-learning model in maritime environments and mitigates image information degradation caused by diverse weather conditions at sea through sensor fusion.

\newpage
\bibliographystyle{IEEEtran}  
\bibliography{main.bib}

\begin{thebibliography}{10}
\providecommand{\url}[1]{#1}
\csname url@samestyle\endcsname
\providecommand{\newblock}{\relax}
\providecommand{\bibinfo}[2]{#2}
\providecommand{\BIBentrySTDinterwordspacing}{\spaceskip=0pt\relax}
\providecommand{\BIBentryALTinterwordstretchfactor}{4}
\providecommand{\BIBentryALTinterwordspacing}{\spaceskip=\fontdimen2\font plus
\BIBentryALTinterwordstretchfactor\fontdimen3\font minus \fontdimen4\font\relax}
\providecommand{\BIBforeignlanguage}[2]{{%
\expandafter\ifx\csname l@#1\endcsname\relax
\typeout{** WARNING: IEEEtran.bst: No hyphenation pattern has been}%
\typeout{** loaded for the language `#1'. Using the pattern for}%
\typeout{** the default language instead.}%
\else
\language=\csname l@#1\endcsname
\fi
#2}}
\providecommand{\BIBdecl}{\relax}
\BIBdecl

\bibitem{prasad2017video}
D.~K. Prasad, D.~Rajan, L.~Rachmawati, E.~Rajabally, and C.~Quek, ``Video processing from electro-optical sensors for object detection and tracking in a maritime environment: A survey,'' \emph{IEEE Transactions on Intelligent Transportation Systems}, vol.~18, no.~8, pp. 1993--2016, 2017.

\bibitem{ribeiro2017data}
R.~Ribeiro, G.~Cruz, J.~Matos, and A.~Bernardino, ``A data set for airborne maritime surveillance environments,'' \emph{IEEE Transactions on Circuits and Systems for Video Technology}, vol.~29, no.~9, pp. 2720--2732, 2017.

\bibitem{shao2018seaships}
Z.~Shao, W.~Wu, Z.~Wang, W.~Du, and C.~Li, ``Seaships: A large-scale precisely annotated dataset for ship detection,'' \emph{IEEE transactions on multimedia}, vol.~20, no.~10, pp. 2593--2604, 2018.

\bibitem{bovcon2019mastr}
B.~Bovcon, J.~Muhovi{\v{c}}, J.~Per{\v{s}}, and M.~Kristan, ``The mastr1325 dataset for training deep usv obstacle detection models,'' in \emph{Int. Conf. Intell. Robots and Systems}.\hskip 1em plus 0.5em minus 0.4em\relax IEEE, 2019, pp. 3431--3438.

\bibitem{zhang2020integrated}
W.~Zhang, X.~He, W.~Li, Z.~Zhang, Y.~Luo, L.~Su, and P.~Wang, ``An integrated ship segmentation method based on discriminator and extractor,'' \emph{Image and Vision Computing}, vol.~93, p. 103824, 2020.

\bibitem{sun2022irdclnet}
Y.~Sun, L.~Su, Y.~Luo, H.~Meng, Z.~Zhang, W.~Zhang, and S.~Yuan, ``Irdclnet: Instance segmentation of ship images based on interference reduction and dynamic contour learning in foggy scenes,'' \emph{IEEE Transactions on Circuits and Systems for Video Technology}, vol.~32, no.~9, pp. 6029--6043, 2022.

\bibitem{hu2021pag}
J.~Hu, X.~Zhi, T.~Shi, W.~Zhang, Y.~Cui, and S.~Zhao, ``Pag-yolo: A portable attention-guided yolo network for small ship detection,'' \emph{Remote Sensing}, vol.~13, no.~16, p. 3059, 2021.

\bibitem{chen2021wodis}
X.~Chen, Y.~Liu, and K.~Achuthan, ``Wodis: Water obstacle detection network based on image segmentation for autonomous surface vehicles in maritime environments,'' \emph{IEEE Transactions on Instrumentation and Measurement}, vol.~70, pp. 1--13, 2021.

\bibitem{kim2022object}
J.-H. Kim, N.~Kim, Y.~W. Park, and C.~S. Won, ``Object detection and classification based on yolo-v5 with improved maritime dataset,'' \emph{Journal of Marine Science and Engineering}, vol.~10, no.~3, p. 377, 2022.

\bibitem{ribeiro2022real}
M.~Ribeiro, B.~Damas, and A.~Bernardino, ``Real-time ship segmentation in maritime surveillance videos using automatically annotated synthetic datasets,'' \emph{Sensors}, vol.~22, no.~21, p. 8090, 2022.

\bibitem{chen2024efficientship}
H.~Chen, J.~Xue, H.~Wen, Y.~Hu, and Y.~Zhang, ``Efficientship: A hybrid deep learning framework for ship detection in the river.'' \emph{CMES-Computer Modeling in Engineering \& Sciences}, vol. 138, no.~1, 2024.

\bibitem{fefilatyev2006horizon_buoy_dataset}
S.~Fefilatyev, V.~Smarodzinava, L.~O. Hall, and D.~B. Goldgof, ``Horizon detection using machine learning techniques,'' in \emph{2006 5th International Conference on Machine Learning and Applications (ICMLA'06)}.\hskip 1em plus 0.5em minus 0.4em\relax IEEE, 2006, pp. 17--21.

\bibitem{KristanCYB2015}
M.~Kristan, V.~S. Kenk, S.~Kovačič, and J.~Perš, ``Fast image-based obstacle detection from unmanned surface vehicles,'' \emph{IEEE transactions on cybernetics}, vol.~46, no.~3, pp. 641--654, 2016.

\bibitem{bovcon2018stereo}
B.~Bovcon, R.~Mandeljc, J.~Per{\v{s}}, and M.~Kristan, ``Stereo obstacle detection for unmanned surface vehicles by imu-assisted semantic segmentation,'' \emph{Robotics and Autonomous Systems}, vol. 104, pp. 1--13, 2018.

\bibitem{smd_prasad2017video}
D.~K. Prasad, D.~Rajan, L.~Rachmawati, E.~Rajabally, and C.~Quek, ``Video processing from electro-optical sensors for object detection and tracking in a maritime environment: a survey,'' \emph{IEEE Transactions on Intelligent Transportation Systems}, vol.~18, no.~8, pp. 1993--2016, 2017.

\bibitem{bovcon2021mods}
B.~Bovcon, J.~Muhovi{\v{c}}, D.~Vranac, D.~Mozeti{\v{c}}, J.~Per{\v{s}}, and M.~Kristan, ``Mods—a usv-oriented object detection and obstacle segmentation benchmark,'' \emph{IEEE Transactions on Intelligent Transportation Systems}, vol.~23, no.~8, pp. 13\,403--13\,418, 2021.

\bibitem{fan2021rethinking}
M.~Fan, S.~Lai, J.~Huang, X.~Wei, Z.~Chai, J.~Luo, and X.~Wei, ``Rethinking bisenet for real-time semantic segmentation,'' in \emph{Proceedings of the IEEE/CVF conference on computer vision and pattern recognition}, 2021, pp. 9716--9725.

\bibitem{yu2018BiSeNet}
C.~Yu, J.~Wang, C.~Peng, C.~Gao, G.~Yu, and N.~Sang, ``Bisenet: Bilateral segmentation network for real-time semantic segmentation,'' in \emph{Proc. European Conf. Computer Vision}, 2018, pp. 325--341.

\bibitem{yu2021BiSeNet}
C.~Yu, C.~Gao, J.~Wang, G.~Yu, C.~Shen, and N.~Sang, ``Bisenet v2: Bilateral network with guided aggregation for real-time semantic segmentation,'' \emph{International Journal of Computer Vision}, vol. 129, pp. 3051--3068, 2021.

\bibitem{xu2023pidnet}
J.~Xu, Z.~Xiong, and S.~P. Bhattacharyya, ``Pidnet: A real-time semantic segmentation network inspired by pid controllers,'' in \emph{Proceedings of the IEEE/CVF conference on computer vision and pattern recognition}, 2023, pp. 19\,529--19\,539.

\bibitem{vaswani2017attention}
A.~Vaswani, N.~Shazeer, N.~Parmar, J.~Uszkoreit, L.~Jones, A.~N. Gomez, {\L}.~Kaiser, and I.~Polosukhin, ``Attention is all you need,'' \emph{Advances in neural information processing systems}, vol.~30, 2017.

\bibitem{dosovitskiy2020image}
A.~Dosovitskiy, L.~Beyer, A.~Kolesnikov, D.~Weissenborn, X.~Zhai, T.~Unterthiner, M.~Dehghani, M.~Minderer, G.~Heigold, S.~Gelly \emph{et~al.}, ``An image is worth 16x16 words: Transformers for image recognition at scale,'' \emph{arXiv preprint arXiv:2010.11929}, 2020.

\bibitem{zheng2021rethinking}
S.~Zheng, J.~Lu, H.~Zhao, X.~Zhu, Z.~Luo, Y.~Wang, Y.~Fu, J.~Feng, T.~Xiang, P.~H. Torr \emph{et~al.}, ``Rethinking semantic segmentation from a sequence-to-sequence perspective with transformers,'' in \emph{Proceedings of the IEEE/CVF conference on computer vision and pattern recognition}, 2021, pp. 6881--6890.

\bibitem{strudel2021segmenter}
R.~Strudel, R.~Garcia, I.~Laptev, and C.~Schmid, ``Segmenter: Transformer for semantic segmentation,'' in \emph{Proceedings of the IEEE/CVF international conference on computer vision}, 2021, pp. 7262--7272.

\bibitem{carion2020end}
N.~Carion, F.~Massa, G.~Synnaeve, N.~Usunier, A.~Kirillov, and S.~Zagoruyko, ``End-to-end object detection with transformers,'' in \emph{European conference on computer vision}.\hskip 1em plus 0.5em minus 0.4em\relax Springer, 2020, pp. 213--229.

\bibitem{wang2021pyramid}
W.~Wang, E.~Xie, X.~Li, D.-P. Fan, K.~Song, D.~Liang, T.~Lu, P.~Luo, and L.~Shao, ``Pyramid vision transformer: A versatile backbone for dense prediction without convolutions,'' in \emph{Proceedings of the IEEE/CVF international conference on computer vision}, 2021, pp. 568--578.

\bibitem{xie2021segformer}
E.~Xie, W.~Wang, Z.~Yu, A.~Anandkumar, J.~M. Alvarez, and P.~Luo, ``Segformer: Simple and efficient design for semantic segmentation with transformers,'' \emph{Advances in neural information processing systems}, vol.~34, pp. 12\,077--12\,090, 2021.

\bibitem{wang2023internimage}
W.~Wang, J.~Dai, Z.~Chen, Z.~Huang, Z.~Li, X.~Zhu, X.~Hu, T.~Lu, L.~Lu, H.~Li \emph{et~al.}, ``Internimage: Exploring large-scale vision foundation models with deformable convolutions,'' in \emph{Proceedings of the IEEE/CVF Conference on Computer Vision and Pattern Recognition}, 2023, pp. 14\,408--14\,419.

\bibitem{dai2017deformable}
J.~Dai, H.~Qi, Y.~Xiong, Y.~Li, G.~Zhang, H.~Hu, and Y.~Wei, ``Deformable convolutional networks,'' in \emph{Proceedings of the IEEE international conference on computer vision}, 2017, pp. 764--773.

\bibitem{zhu2019deformable}
X.~Zhu, H.~Hu, S.~Lin, and J.~Dai, ``Deformable convnets v2: More deformable, better results,'' in \emph{Proceedings of the IEEE/CVF conference on computer vision and pattern recognition}, 2019, pp. 9308--9316.

\bibitem{lin2014coco}
T.-Y. Lin, M.~Maire, S.~Belongie, J.~Hays, P.~Perona, D.~Ramanan, P.~Doll{\'a}r, and C.~L. Zitnick, ``Microsoft coco: Common objects in context,'' in \emph{European conference on computer vision}.\hskip 1em plus 0.5em minus 0.4em\relax Springer, 2014, pp. 740--755.

\bibitem{pascal-voc-2012-xkk13_dataset}
\BIBentryALTinterwordspacing
PASCAL, ``Pascal voc 2012 dataset,'' \url{ https://universe.roboflow.com/mati-frenkel/pascal-voc-2012-xkk13 }, aug 2021, visited on 2024-05-27. [Online]. Available: \url{https://universe.roboflow.com/mati-frenkel/pascal-voc-2012-xkk13}
\BIBentrySTDinterwordspacing

\bibitem{song2023extreme}
D.~Song, Z.~Chen, Y.~P. Wang, and Y.~Wu, ``Extreme weather events induced coastal environment changes under multiple anthropogenic impacts,'' p. 1214718, 2023.

\bibitem{fischer2015anthropogenic}
E.~M. Fischer and R.~Knutti, ``Anthropogenic contribution to global occurrence of heavy-precipitation and high-temperature extremes,'' \emph{Nature climate change}, vol.~5, no.~6, pp. 560--564, 2015.

\bibitem{chen2024weather}
M.~Chen, J.~Sun, K.~Aida, and A.~Takefusa, ``Weather-aware object detection method for maritime surveillance systems,'' \emph{Future Generation Computer Systems}, vol. 151, pp. 111--123, 2024.

\bibitem{howard2019searching}
A.~Howard, M.~Sandler, G.~Chu, L.-C. Chen, B.~Chen, M.~Tan, W.~Wang, Y.~Zhu, R.~Pang, V.~Vasudevan \emph{et~al.}, ``Searching for mobilenetv3,'' in \emph{Proceedings of the IEEE/CVF international conference on computer vision}, 2019, pp. 1314--1324.

\bibitem{cai2018proxylessnas}
H.~Cai, L.~Zhu, and S.~Han, ``Proxylessnas: Direct neural architecture search on target task and hardware,'' \emph{arXiv preprint arXiv:1812.00332}, 2018.

\bibitem{liu2019variance}
L.~Liu, H.~Jiang, P.~He, W.~Chen, X.~Liu, J.~Gao, and J.~Han, ``On the variance of the adaptive learning rate and beyond,'' \emph{arXiv preprint arXiv:1908.03265}, 2019.

\end{thebibliography}


\end{document}